\documentclass{article}
\usepackage{ijcai22}

\usepackage{times}

\usepackage{soul}
\usepackage{url}
\usepackage{hyperref}
\usepackage[utf8]{inputenc}
\usepackage[small]{caption}
\usepackage{graphicx}
\usepackage{amsmath}
\usepackage{amssymb}
\usepackage{bm}
\usepackage{color}
\usepackage{booktabs}
\usepackage{subcaption}
\usepackage{xr}
\usepackage{float}

\usepackage[dvipsnames]{xcolor}
\DeclareMathOperator{\EX}{\mathbb{E}} 

\newcommand{\bx}{\bm{x}}
\newcommand{\br}{\bm{r}}
\newcommand{\by}{\bm{y}}
\newcommand{\bz}{\bm{z}}

\newcommand{\bd}{\bm{d}}

\title{Variational Learning for Unsupervised Knowledge Grounded Dialogs}

\author{
Mayank Mishra\footnote{Contact Author}\and
Dhiraj Madan\and
Gaurav Pandey\And
Danish Contractor\\
\affiliations
IBM Research AI\\
\emails
mayank.mishra1@ibm.com,
\{dmadan07, gpandey1\}@in.ibm.com,
danish.contractor@ibm.com
}

\begin{document}

\maketitle

\begin{abstract}
Recent methods for knowledge grounded dialogs generate responses by incorporating information from an external textual document \cite{RAG,Guu20}. These methods do not require the exact document to be known during training and rely on the use of a retrieval system to fetch relevant documents from a large index. The documents used to generate the responses are modeled as latent variables whose prior probabilities need to be estimated. Models such as RAG \cite{RAG} and REALM \cite{Guu20}, marginalize the document probabilities over the documents retrieved from the index to define the log likelihood loss function which is optimized end-to-end. 

In this paper, we develop a variational approach to the above technique wherein, we instead maximize the Evidence Lower bound (ELBO). Using a collection of three publicly available open-conversation datasets, we demonstrate how the posterior distribution, that has information from the ground-truth response, allows for a better approximation of the objective function during training.  To overcome the challenges associated with sampling over a large knowledge collection, we develop an efficient approach to approximate the ELBO. To the best of our knowledge we are the first to apply variational training for open-scale unsupervised knowledge grounded dialog systems.
\end{abstract}

\section{Introduction}

\begin{figure*}[ht]
    \centering
    \vspace{-3ex}
    \includegraphics[scale=0.6]{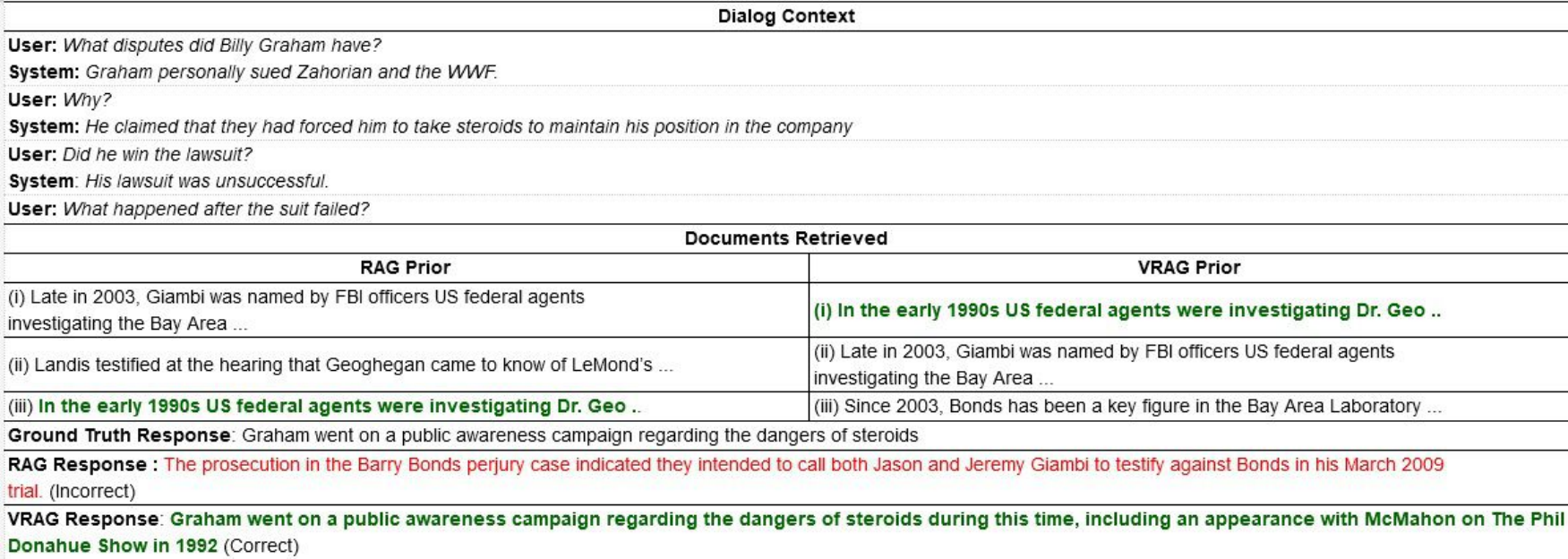}
    \caption{Documents from the OR-QuAC dataset \protect\cite{OpenQuAC} retrieved by RAG \protect\cite{RAG} and VRAG (our approach), shown in decreasing order of probabilities, along with the response generated by each model. The document highlighted in green is the correct document required to generate the response. Document text truncated for ease of presentation.}
    \label{fig:example}
\end{figure*}

In this paper we focus our attention on the task of generating responses, grounded on information present in a large collection of external textual documents \cite{RAG}. In real-world scenarios, the exact document that one must access for generating the response is often unknown and one only has access to conversation logs and a document collection. Hence during training, given a dialog context, the primary challenge is first figuring out the correct document needed to generate the response, and then using that document for generating the actual response. Figure \ref{fig:example} shows an example dialog from the OR-QuAC dataset \cite{OpenQuAC}. Here, the user is asking the system about what happened when a lawsuit filed by Billy Graham\footnote{\url{https://en.wikipedia.org/wiki/Superstar\_Billy\_Graham}} failed. Responding to this question in the dialog context, without relying on external knowledge isn't possible. The figure shows the ground truth response along with the correct document needed to generate the response.

A straightforward baseline approach would be to use an out-of-the-box retriever (for instance, a tf-idf based retriever such as BM25~\cite{BM25} or a neural retriever such as DPR~\cite{dpr20}) for first retrieving the document and then using a retrieved document for generating the response. While this is fairly easy to implement, it cannot be trained in an end-to-end manner and thus, the retriever never improves as the model learns to generate responses. 

\begin{figure}[t]
    \centering
    \vspace{-2ex}
    \begin{subfigure}[b]{0.35\textwidth}
        \centering
        \includegraphics[width=\textwidth, scale=1]{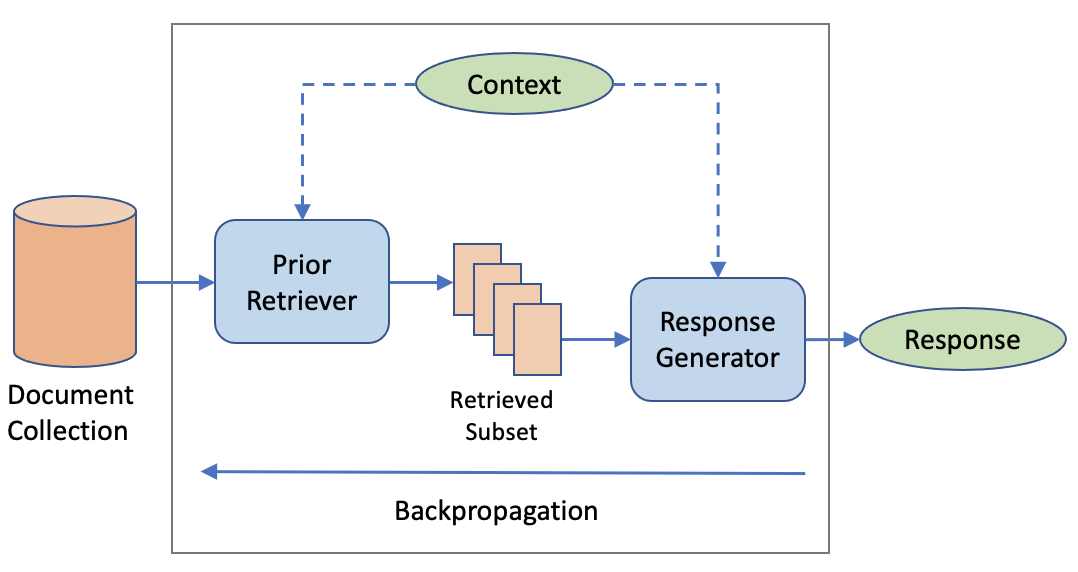}
        \caption{RAG \protect\cite{RAG}}
        \label{fig:RAG}
    \end{subfigure}
    \begin{subfigure}[b]{0.35\textwidth}
        \centering
        \includegraphics[width=\textwidth, scale=1]{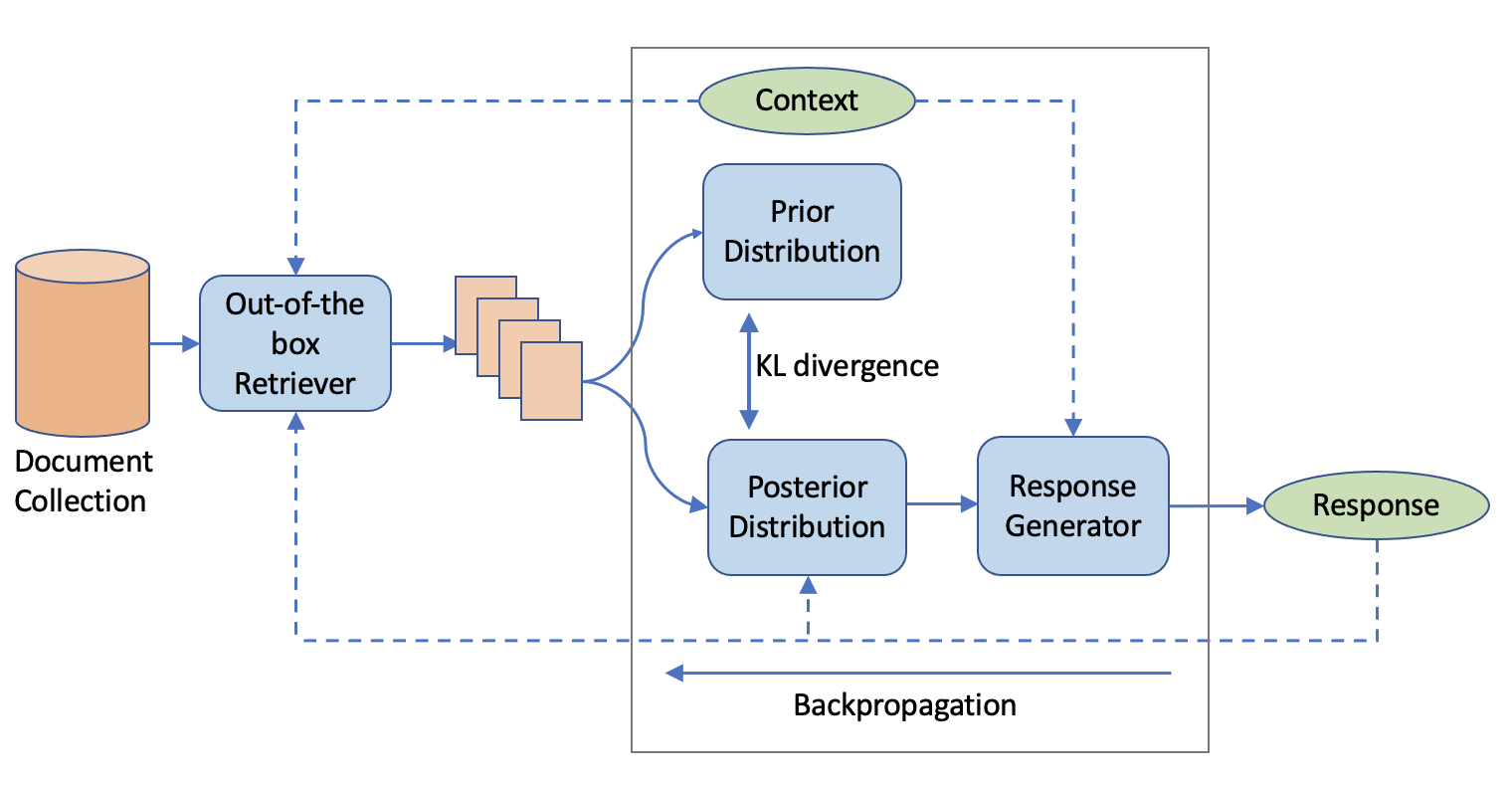}
        \caption{Variational models with pre-retrieval}
        \label{fig:VM}
    \end{subfigure}
    \begin{subfigure}[b]{0.35\textwidth}
        \centering
        \includegraphics[width=\textwidth, scale=1]{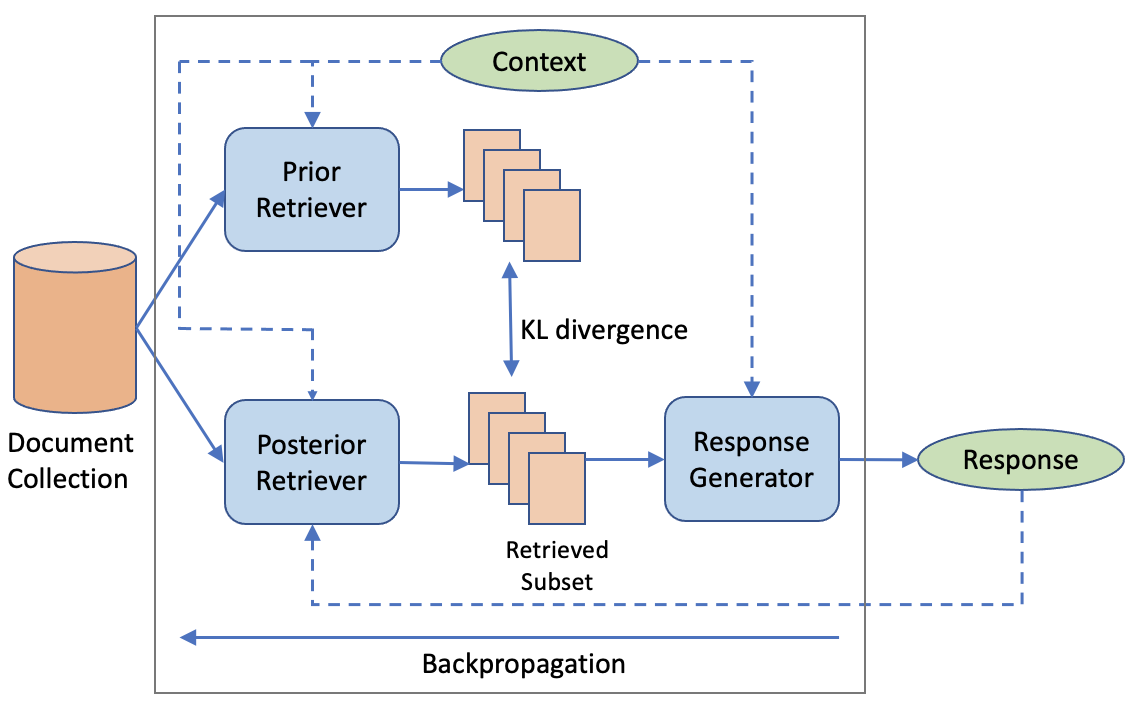}
        \caption{VRAG (Our approach)}
        \label{fig:VRAG}
    \end{subfigure}
    \caption{(a) RAG does not use information from the responses to retrieve documents (b) Existing variational methods \protect\cite{Lia19,neurips_lin} use pre-retrieval (retriever not updated during training) to get a small set of documents to work with. (c) Our approach - VRAG which is trained end-to-end and uses the response to train a posterior distribution which guides the prior distribution}
    \label{fig:models}
\end{figure}

To overcome this limitation, methods such as RAG~\cite{RAG}, model documents as latent variables and learn a distribution over these variables (Figure \ref{fig:RAG}). This distribution is referred to as the {\em document prior}. Specifically, the document-prior distribution is defined by querying a knowledge index \cite{JDH17,dpr20} using the dialog context (history), and then converting the retrieval scores of the top-k\footnote{typically $k$ $=$ 5-10.} documents into a probability distribution. The response-likelihood can be defined using any neural language generator such as GPT2 \cite{GPT2}. It then performs a marginalisation of the latent variable over the retrieved documents to compute the approximate probability of the response, given the context. The negative log likelihood under this approximation forms the loss function to train.

However, one of the weaknesses of this approach is that, using the document-prior to query the index during training, ignores crucial information present in the ground-truth response which could have aided document retrieval. As a result, the response-likelihood network parameters may receive a weaker signal during training, which, in-turn, can cause models to try reducing their dependence on external knowledge by `memorizing', especially if the correct document is rarely fetched by the retriever.

\noindent {\bf Variational Retrieval-Augmented Generation (VRAG): }In this paper, we propose an approach that overcomes this limitation. We incorporate the ground truth response with the dialog context for retrieving the documents during training in a secondary retriever. This increases the chances of retrieving the correct document during training. The distribution over the documents defined by this retriever is referred to as the {\em document posterior}. The document posterior guides the training of the document prior while the documents sampled by the posterior are fed to the decoder for generating the response. Such a formalism emerges naturally in the variational setting, wherein the evidence lower bound (ELBO) is optimized instead of the maximum likelihood objective. Hence, we refer to the model as Variational Retrieval-Augmented Generation (VRAG).

One of the advantages of variational training is that it provides a low variance estimate of the objective (as compared to sampling from the document-prior distribution), for the same number of samples. Although this has been used in supervised settings (\cite{Che20} and \cite{KAK20}), we note that directly training under the variational objective may be prohibitively expensive in case of a very large\footnote{Collections can have millions of documents} document collection (sampling an element from the posterior distribution, would require retrieval scores for each document in the index collection). In fact, related approaches for variational training therefore only use a small set of pre-retrieved documents~\cite{Lia19,neurips_lin} to overcome this bottleneck. In particular, such approaches first use an out-of-the box retriever to fetch a small set of documents (typically 5-10 documents) from the entire document collection. The methods then learn a prior as well as posterior distribution over the small set of pre-retrieved documents only by optimizing the variational objective (Figure \ref{fig:VM}).

A major weakness of this approach is that the out-of-the-box retriever does not benefit from training. As a result, if the recall of the out-of-the box retriever is low, that is, the correct document is not present in the pre-retrieved subset for most of the training data, the mapping from the document to the response learnt will be highly noisy and not very useful (we also demonstrate this experimentally in this paper).

\noindent {\bf Contributions:} In this paper we describe our approach called VRAG or Variational Retrieval-Augmented Generation\footnote{We provide the code and the supplementary material at \url{https://github.com/mayank31398/VRAG} and \url{https://arxiv.org/abs/2112.00653} respectively.} which allows us to extend variational optimization to cases where documents are retrieved from large document collections. Instead of pre-retrieving a small set of documents to facilitate variational training, we retrieve documents from the entire collection but perform a summation over the top-k retrieved documents from the posterior distribution as well as the prior distribution to approximate the variational objective (Figure \ref{fig:VRAG}). Top-k retrieval can be performed efficiently using an index for nearest neighbor search such as Faiss \cite{JDH17}, and we find that this simple trick performs significantly better than other approaches.
We present experiments on three, publicly available, conversational QA datasets and we show that variational training helps build better knowledge grounded dialog systems. 
Our experiments show that not only does VRAG perform better on the end-task, it also learns a better retriever.
To the best of our knowledge, we are the first to apply variational training for open-scale unsupervised knowledge grounded dialog systems.\footnote{Concurrently, \cite{paranjape2022hindsight} also use variational training with RAG to generate responses}

\section{Background}
As is commonly done in dialog modeling tasks, we represent the collection of dialogs as a set of context (dialog history) and response pairs;  $\mathcal{T}=\{(\bx^{(i)},\by^{(i)})\}_{i=1}^m$ where each context $\bx^{(i)}$ as well as its response $\by^{(i)}$ is a sequence of tokens. Further let $\mathcal{D}=\{\bd_j\}_{j=1}^N$ be a set of documents in the form of a large indexed document collection.
We assume that each context-response pair requires exactly 1 document $\bd_j \in \mathcal{D}$ (where $1 \leq j \leq N$) to generate the corresponding response. Let $\bz^{(i)}$ denote a discrete variable which indicates the document (from the indexed collection) needed for training instance $i$ i.e,  $\bd_{\bz^{i}} \in \mathcal{D}$. We can now model the joint likelihood of a response and document pair $(\by^{(i)}, \bz^{(i)})$ as $ p(\by^{(i)}, \bz^{(i)} \vert \bx^{(i)}) = p(\bz^{(i)}\vert \bx^{(i)}) p(\by^{(i)} \vert \bx^{(i)}, \bz^{(i)})$.

In the absence of document-level supervision, the $\bz^{(i)}$ variables are unknown or `latent'. Here we will be maximizing $\log p(\by^{(i)} \vert \bx^{(i)})= \sum_{\bz} p(\bz^{(i)} \vert \bx^{(i)}) p(\by^{(i)} \vert \bz^{(i)},\bx^{(i)})$. However, since an explicit summation over the entire document collection can be computationally intractable, one needs to resort to a few approximation techniques. For ease of notation, we will drop the superscript $(i)$ from now on.

\noindent \textbf{Retrieval based approaches:} Approaches such as RAG~\cite{RAG} and REALM ~\cite{Guu20}, maintain an index which allows one to retrieve the top-k documents with high prior probability. The objective is then approximated as a sum over these retrieved documents.
    
Specifically, the document-prior distribution $p(\bz \vert \bx)$ is defined based on scores returned by the Dense Passage Retriever (DPR) \cite{dpr20}.

The top-k most relevant documents ($S^p_k)$ for a query (dialog context) are retrieved from an index that allows efficient retrieval \cite{JDH17} using MIPS search. We denote the approximate document-prior distribution, normalized over the set $S^p_k$, as $\hat{p}(\bz \vert \bx)$. The overall objective for generating the response can then be written as $\log \left[ \sum_{\bz \in S^p_k} \hat{p}(\bz \vert \bx) p (\by \vert \bz,\bx) \right
]$. RAG suffers from a drawback that it does not use the information from responses in order to retrieve documents for a given training instance.

\noindent \textbf{Variational techniques:} An alternative approach is to maximize a variational lower bound on the objective. Here we need to define an Evidence Lower Bound (ELBO) on the likelihood as $\log p(\by \vert \bx) \geq \EX_{\bz \sim q(\bz \vert \bx,\by)} \left[ log \left( \frac{p(\by,\bz\vert \bx)}{q(\bz \vert \bx,\by)} \right) \right]$. This lower bound holds for any distribution $q$. Variational autoencoders \cite{KW13}  define another network to model the distribution $q$. To train such networks, the ELBO is split as:
\begin{equation}
\label{eq:elbo}
\EX_{\bz\sim q} \left[ \log p(\by\vert \bz,\bx) \right] - KL \left[ q \|p(\bz \vert \bx) \right]
\end{equation}

The first term is an expectation that can be estimated by sampling documents from the document-posterior $q(\bz\vert \bx,\by)$ distribution. The response-likelihood network is then run only using these sampled documents. One can either use re-parameterization trick (with Gumbel softmax distribution) \cite{Gumbel,Concrete} or policy gradient method to back propagate through the sampling step. However, in order to sample a document, one would need to compute the entire distribution over the documents. This can be prohibitively expensive when using a large document collection. 
The second term (KL-divergence) is also computed as an explicit sum, given access to prior and posterior probability distributions, and is also intractable for large document collections.

In summary, variational training which uses the posterior distribution to retrieve documents while training,  can help retrieve more relevant documents for training the response-generator. One of the trivial ways to extend these approaches (for large document corpus) is to identify the candidate knowledge documents for each training instance via an existing out-of-the-box retriever (\cite{neurips_lin}, \cite{Lia19}). One can then create prior and posterior distributions on the restricted set of documents.  However this does not allow us to train the retriever and we also show in our experiments, that the trained distribution in such a setting does a poor job of generalizing as a retriever (Section \ref{sec:expts}). 

In our approach we generalize the variational technique to open domain setting without fixing or pre-retrieving those candidate documents. In order to do so, we would first need to be able to compute the ELBO objective more efficiently (over the entire document corpus) as described in the next section.

\section{Variational RAG (VRAG)}
\label{sec:VRAG}
Variational training involves using both, the document-prior ($p(\bz \vert \bx)$) and document-posterior distributions ($q(\bz \vert \bx, \by)$). We model each of these based on scores from a Dense Passage retriever (DPR) \cite{dpr20} i.e.
\begin{align}
    \label{eq:vragprior}
    p(\bz\vert \bx)&= \text{softmax}\left( f(\bz)^T g(\bx) \right)\\
    \text{ and }\label{eq:vragpost}q(\bz\vert \bx, \by)&= \text{softmax}\left( f(\bz)^T h(\bx, \by) \right)
\end{align}
where $f$ and $g$ are parameterized representations of documents ($\bz$) and dialog contexts ($\bx$).  $h(\bx,\by)$ denotes the joint embedding of the context-response pair. These are created using neural models such as BERT \cite{BERT}. We use a single network to compute the document embeddings $f(\bz)$ for both prior and posterior.

In order to efficiently compute expectation and KL divergence terms in the ELBO objective (Equation \ref{eq:elbo}), we need to approximate the above distributions. To do so, we maintain an index on document embeddings. This allows us to retrieve the set of top-k documents under prior and posterior distributions (equations \ref{eq:vragprior} and \ref{eq:vragpost}) using MIPS search \cite{JDH17}. Note that since the query embeddings $g$ and $h$ are trainable, the retriever is trained over the epochs as well. We denote the sets of top documents  under prior and posterior distributions by $S^p_k$ and $S^q_k$ respectively.  

The overall cost is then computed using Evidence Lower Bound (Equation \ref{eq:elbo}). Here for the first term, we normalize $\hat{q}$ over the set of top-k documents (denoted by $S_k^q$) returned by the index when queried using the posterior (i.e, the response and the dialog context). The first term is then approximated as $\sum_{\bz \in S^q_k} \hat{q}(\bz \vert \bx,\by ) \log {p} (\by \vert \bz,\bx)$. To approximate the KL-divergence we use the top-k knowledge instances retrieved by querying the index using the dialog-context (for prior) and the context-response pair (for posterior). We then take union of the two sets  $S^p_k$ and $S^q_k$ to form the set $S_{KL}$. We use this set ($S_{KL}$) to approximate the KL-divergence. Thus, the KL-divergence in Equation \ref{eq:elbo} is given by:
\begin{equation}
KL \left[ \hat{q} \vert \vert \hat{p} \right] =\sum_{\bz \in S_{KL}} \hat{q}(\bz) \log \left( \frac{\hat{q}(\bz)}{\hat{p}(\bz)} \right),
\end{equation}
where the approximate posterior ($\hat{q}$) and prior ($\hat{p}$) in this case are obtained by normalizing the retrieval scores on $S_{KL}$. The intuition behind this approximation is that the documents with low posterior probability do not contribute much to the KL objective and hence, can safely be ignored. Similar to other variational models, VRAG is trained end-to-end.

\subsection{Architecture}
We now describe the neural networks used to model the prior and posterior distributions, and the response generator. 

\noindent{\bf Prior Distribution Encoders:} We need to create document and context representations defined by functions $f$and $g$, respectively for modelling the prior. 
We pass the input context through BERT model \cite{BERT} to create context representation. We use special markers to separate the turns. The embedding at final layer of $[CLS]$ token is passed through a linear layer to create context representation.
Similarly the document is passed through (separate) BERT Model to create a document representation at $[CLS]$ token.

\noindent{\bf Posterior Distribution Encoders:} Similar to the modeling of the prior distribution, we use another (separate) BERT to model the posterior. Here we create the input representation of the context-response  pair ($\bx$, $\by$), where a special marker is used to separate the context and response.

\noindent{\bf Response Likelihood:} We use the standard sequence to sequence formulation for response likelihood, where
$\log p(\by\vert \bz,\bx) = \sum_j \log p(\by_j\vert \by_{<j},\bz,\bx)$.
Here, we use the GPT2~\cite{GPT2} model as our decoder with input sequence consisting of context and response.

\subsection{Training details}
We train our network to maximize the ELBO objective (Equation \ref{eq:elbo}). We initialize our document-prior (for both RAG and VRAG) and document-posterior (for VRAG) networks with the pretrained DPR-Multiset model\footnote{This can be used to initialize a model in Hugging Face, see \url{https://huggingface.co/transformers/model_doc/dpr.html}} pre-trained using data from the Natural Questions \cite{NQ}, TriviaQA \cite{TriviQA}, WebQuestions \cite{WebQuestions} and CuratedTREC \cite{CuratedTREC} datasets.

During the training of the model, it can be difficult to re-build the document index after every change to the document representation parameters in $f$, therefore similar to \citeauthor{RAG}, the parameters in $f$ are kept constant. 

We used early stopping with $patience=5$ on recall of the validation sets to prevent overfitting of models. 
The loss was optimized using AdamW Optimizer \cite{AdamW}. We also found it useful to continue training the response-likelihood for both RAG and VRAG after the joint training is complete. This is because while training, the decoder-likelihood function often lags behind prior (for RAG) and posterior (for VRAG).

\subsection{Response Generation} At test time we need to generate the response for a given context by first retrieving a document -- thus, in both RAG and VRAG models we use the trained document-prior model to retrieve the top-k documents using the dialog context as query.  We experiment with two different decoding strategies to generate the response:
\begin{enumerate}
  \item \textbf{Top Document Decoding: } In this case the document with the highest prior probability is used to condition the generator. The response is then generated using beam search (beam width=3) on the trained GPT-2 response generation model; the most likely beam is taken to be the prediction of the model. We refer to this method as the `top-1 decoding' in our experiments.
  \item \textbf{Top-k Documents Decoding: } Here top (k=5) documents are retrieved from the prior distribution, say $\bz_1,...,\bz_5$. A beam search is then run to generate the top response from each of these say $\br_1,...,\br_5$. We use the estimate of $p(\br_i \vert \bx) \approx p(\bz\vert \bx) p(\br_i \vert \bz,\bx)$.
  The most likely response under the estimated distribution is taken to be the response generated by the model\footnote{This is the same as ``Fast Decoding'' as defined in \cite{RAG}.}. We refer to this as `top-5 decoding' in our experiments.
\end{enumerate}

\section{Experiments} \label{sec:expts}
Our experiments aim to answer the following questions: (1)  Does Variational RAG (VRAG), which uses samples from the approximated document-posterior distribution, perform better than vanilla RAG? (2)  Does the quality of generated responses improve by decoding using the top-k documents? (3) How do the trained document retriever modules for RAG and VRAG compare with each other? (4) Are the quality of samples returned by the document-posterior of VRAG better than document-prior of RAG as hypothesized? (5) How does VRAG compare with other approximations for variational training?

\subsection{Datasets}
\noindent{\bf OR-QuAC}~\cite{OpenQuAC}: This dataset is a modified version of the QuAC \cite{QuAC} dataset. The dataset consists of dialog conversations, where each conversation is associated with the top-5 most relevant documents (retrieved using TF-IDF \cite{BM25} based BM25 ranking) from the QuAC dataset. To create an open-scale collection for our task, we index the set of all the documents available in the train, validation and test splits. In some cases of the test and validation set, the ground-truth document may be missing in the top-5 list associated with each conversation. In such cases, we obtain the ground-truth document from the original QuAC dataset and add it to the indexed collection.

\noindent{\bf DSTC9}~\cite{DSTC9}: This dialog dataset was released as part of the DSTC9 challenge. The dataset comprises of dialog conversation turns in which the system: (i) needs to identify the turns in which to consult a collection for textual FAQs, (ii) retrieve the FAQ if required, (iii) and then generate the response based on the retrieved FAQ. The universe of knowledge documents in this dataset is the set of FAQs and each FAQ also includes the entity name because the same question can occur multiple times for different entities (eg: ``Is parking available?"). The training dataset consists of conversations based on 4 different domains i.e hotels, restaurants, trains and taxis. The test dataset contains an additional domain, attractions, which is not found in the train and validation splits. 

\noindent{\bf DoQA}~\cite{DoQA}: This dataset comprises of open-ended dialog conversations on different domains like cooking, travel and movies. Unlike, the OR-QuAC and DSTC9 datasets, most questions in this dataset are not factoid/specific questions, and are open-ended. We only use the cooking split for both training and testing.

We preprocess all the datasets by removing all the examples if the ground truth response is ``CANNOTANSWER'' (unanswerable). Each `instance' refers to a context-response pair. 

\noindent {\bf Question-Answering Task}: The OR-QuAC dataset also contains non-contextual variations of questions at each dialog turn and we use them in a QA setting (referred to as {OR-QuAC-QA}).

\begin{table}
\scriptsize
\begin{center}
\begin{tabular}{|c|c|c|c|c|}
\hline

{\bf Dataset} & {\bf Model} & R@1 & R@5 & MRR@5
\\\hline
& DPR & 19.8 & 49.0 & 0.315
\\
OR-QuAC & pre-VM & 17.71 & 38.61 & 0.264
\\
\cite{OpenQuAC} & RAG &  24.2 & 54.9 & 0.366
\\
& VRAG & \textbf{26.0} & \textbf{56.9} & \textbf{0.388}
\\
\hline
& DPR & 13.2 & 34.3 & 0.208
\\
DSTC9 & pre-VM & 28.32 & 42.76 & 0.339
\\
\cite{DSTC9} & RAG & 69.4 & 84.3 & 0.758
\\
& VRAG & \textbf{73.3} & \textbf{87.0} & \textbf{0.792}
\\
\hline
& DPR & 54.3 & 71.7 & 0.612
\\
DoQA & pre-VM & 60.81 & 77.43 & 0.672
\\         
\cite{DoQA} & RAG & 61.8 & 79.0 & 0.687
\\
& VRAG & \textbf{62.4} & \textbf{79.1} & \textbf{0.691}
\\
\hline
& DPR & 9.6 & 30.7 & 0.174
\\
OR-QuAC-QA & pre-VM & 6.27 & 17.26 & 0.104
\\
\cite{OpenQuAC} & RAG & 15.0 & 38.1 & 0.239
\\
& VRAG & \textbf{16.4} & \textbf{41.0} & \textbf{0.261}
\\
\hline

\end{tabular}
\caption{Comparison of VRAG and RAG models in terms of retrieval accuracy (document recall $R@K$ and MRR scores)  on all datasets.}
\label{table:recall-all}
\end{center}
\end{table}

\subsection{Baselines}
\label{sec:baselines}
Apart from our approach (VRAG), we also study the performance of RAG, as well as, a pipeline model which uses the pre-trained DPR-Multiset Retriever and a GPT2 based decoder which is fine-tuned to generate responses. We refer to this as the DPR + GPT2 baseline in our experiments. In addition, we also show the importance of using a trainable retriever for variational setting by comparing gainst a variational model where a fixed set of candidate documents are retrieved using a pre-trained DPR as a retriever (conditioned on context and response) during training \cite{Lia19,neurips_lin}. This model retrieves from the entire document collection using the prior distribution during inference. We refer to this baseline as Variational Model with pre-retrieval (``pre-VM''). Thus, the distributions being trained do not change the set of candidate documents used as training progresses. All models use their respective document-prior distributions during testing.

\subsection{Evaluation metrics}
For each of our experimental runs, we report the Mean Reciprocal Rank@5 (MRR@5), Recall@1 (R@1), Recall@5 (R@5) to evaluate prior's performance. We also report BLEU scores (both with top-1 and top-5 decoding) to assess the performance of generator. The BLEU-1 and BLEU-4 scores in our tables are denoted by B-1 and B-4 respectively. We also consider ``BLEU-penalized'' scores (indicated by BP-1 anad BP-4) which consider the BLEU score at a given test instance as 0 if the document retrieved for the given instance is incorrect. These help ensure that a model is not able to produce a high score by memorizing on a particular given domain.

\begin{table*}[ht]
\scriptsize
\vspace{-2ex}
\begin{center}
\begin{tabular}{|c|c|c|c|c|c|c|c|c|c|}
\hline
& & \multicolumn{4}{c|}{top-1 decoding} & \multicolumn{4}{c|}{top-5 decoding}
\\\hline
{\bf Dataset} & {\bf Model} & {\bf B-1} & {\bf B-4} & {\bf BP-1} & {\bf BP-4} & {\bf B-1} & {\bf B-4} & {\bf BP-1} & {\bf BP-4}
\\
\hline
& DPR + GPT2 & 13.65 & 6.11 & 4.41 & 3.11 & 16.06 & 7.97 & 11.36 & 7.37
\\
OR-QuAC & RAG & 12.88 & 5.94 & 4.60 & 3.03 & 15.39 & 7.64 & 11.72 & 7.21
\\
\cite{OpenQuAC} & pre-VM & 11.44 & 4.87 & 3.52 & 2.36 & 13.55 & 6.26 & 9.17 & 5.89
\\
& VRAG & \textbf{13.97} & \textbf{7.58} & \textbf{5.61} & \textbf{4.02} & \textbf{16.30} & \textbf{9.11} & \textbf{13.10} & \textbf{8.72}
\\
\hline
& DPR + GPT2 & 31.84 & 7.21 & 4.37 & 1.08 & 31.81 & 7.17 & 11.14 & 2.60
\\
DSTC9 & RAG & 33.28 & 8.26 & 25.87 & 6.86 & 33.30 & 8.27 & 28.75 & 7.45
\\
\cite{DSTC9} & pre-VM & 31.57 & 7.16 & 9.92 & 2.66 & 31.87 & 7.29 & 14.7 & 3.98
\\
& VRAG & \textbf{33.49} & \textbf{8.70} & \textbf{26.49} & \textbf{7.57} & \textbf{33.51} & \textbf{8.67} & \textbf{29.80} & \textbf{8.03}
\\
\hline
& DPR + GPT2 & 21.26 & 14.31 & 17.83 & 14.20 & 22.60 & 15.73 & 20.53 & 15.62
\\
DoQA & RAG & \textbf{23.59} & \textbf{17.04} & 20.86 & 16.92 & \textbf{24.27} & \textbf{17.73} & \textbf{22.75} & \textbf{17.60}
\\
\cite{DoQA} & pre-VM & 23.33 & 16.70 & 20.47 & 16.5 & 23.79 & 17.21 & 22.24 & 17.07
\\
& VRAG & 23.38 & 17.02 & \textbf{20.91} & \textbf{16.94} & 23.29 & 16.88 & 21.93 & 16.80
\\
\hline
& DPR + GPT2 & 9.11 & 1.88 & 1.81 & 1.07 & 10.18 & 2.61 & 4.95 & 2.38
\\
OR-QuAC-QA & RAG & 9.12 & 2.31 & 2.36 & 1.44 & 10.39 & 2.97 & 5.75 & 2.75
\\
\cite{OpenQuAC} & pre-VM & 6.96 & 1.17 & 1.11 & 0.67 & 7.84 & 1.75 & 3.05 & 1.67
\\
& VRAG & \textbf{9.64} & \textbf{2.93} & \textbf{2.83} & \textbf{1.66} & \textbf{10.65} & \textbf{3.49} & \textbf{6.73} & \textbf{3.36}
\\
\hline
\end{tabular}
\caption{Comparison of VRAG and RAG models in terms of BLEU and BLEU-penalized on all datasets after decoder fine-tuning.}
\label{table:all}
\end{center}
\end{table*}
\begin{table}[ht]
\vspace{-1ex}
\centering
\scriptsize
\begin{tabular}{|c|c|c|}
\hline
& B-1 (top-1  decoding) & B-1 (top-5 decoding)
\\
\hline
DPR & -42.28\% & -48.64\%
\\
RAG & -36.77\% & -45.19\%
\\
pre-VM & -41.89\% & -44.20\%
\\
VRAG & -43.22\% & -49.88\%
\\
\hline
\end{tabular}
\caption{Percentage drop in BLEU score when correct documents have been removed on OR-QuAC \protect\cite{OpenQuAC} dataset.}
\label{tab:memorization}
\end{table}

\subsection{Results}
As can be seen in Table \ref{table:recall-all}, VRAG outperforms RAG on each dataset for document retrieval. We also note that both RAG and VRAG significantly improve the performance of the initial DPR based retriever. Table \ref{table:all}  shows the performance of the models on the response generation task. Note that all results in Table \ref{table:all} are obtained after further fine tuning the generator networks (after joint training is complete). The results show that the VRAG model outperforms the RAG model on language generation tasks (BLEU metrics) on all datasets except DoQA.  We believe this difference is due to the nature of documents used in this dataset - other datasets have a lot more fact based questions to be answered using knowledge in documents, while more than 66\% of the questions in the DoQA dataset are non-fact based and open-ended.

In addition, it is also possible that the RAG model is memorizing and overfitting on this dataset. For instance, see gains in penalized BLEU scores (BP-1 and BP-4) of RAG and VRAG over DPR in OR-QuAC and DSTC9 datasets in Table \ref{table:all} -- the relative gain of VRAG (over RAG) is significantly higher. This suggests that VRAG model is more likely to generate the response using the correct document and not by merely memorize on a given domain.

\noindent{\bf Alternative approximations for Variational Training: } In Table \ref{table:recall-all} we see that the recall scores of pre-VM model are much worse than our VRAG model.  This observation validates our hypothesis that, because the retrieved samples are not improved during training, the prior and posterior distributions in pre-VM end up focusing only on the few (potentially incorrect) initially-retrieved documents. This is especially problematic if the initial recall is low.  As training progresses, this may worsen the distributions over the initial model (eg: see DSTC9 recall scores for DPR and pre-VM in Table \ref{table:recall-all}) because the correct document isn't present in the retrieved sample, thus giving it an incorrect signal.

\noindent{\bf Effect of Top-5 Decoding:} From Table \ref{table:all} we find that in almost all cases, using top-5 decoding to generate responses performs better than using the single (best scored) document to generate responses. This indicates that models are to able to incorporate information from the correct document even if it is not returned as the top-ranked document.

\noindent{\bf Benefit of Document Posterior:} In Section \ref{sec:VRAG}, we motivated the VRAG model by suggesting that using the posterior to sample documents while training the decoder could help train a better model. We find that recall of the document posterior in VRAG is nearly 14-70\% higher than the document prior of RAG (depending on the dataset). Further, we find that the use of the responses by VRAG (posterior), to query the index during training, results in significantly better retrieval accuracy than RAG (prior) at every epoch during training. While this is perhaps intuitive and expected, our results on recall in Table \ref{table:all} demonstrate the VRAG (prior) which is trained indirectly via the KL-divergence with VRAG (posterior) also outperforms the RAG (prior). We attribute this gain to the fact that the generator in VRAG learns to focus well on the generated documents which further reinforces the posterior network (and indirectly the prior network through KL term) to improve.

\noindent{\bf Study of Memorization:}
One of the issues in such an unsupervised learning is that generator may fail to use the retrieved knowledge. Instead the parameters might have been trained to internally model the external knowledge sources themselves, refered to as `memorization'.
In order to study memorization in the models, we compared the results on response generation of all models in the absence of the correct document -- if the correct document is missing, the models should perform very poorly. A good performance even without obtaining correct document, would indicate lesser reliance on external knowledge and hence a higher tendency for `memorization'.  We, thus rebuild the document index without including any of the documents from the test set and then re-evaluate the performance of our models. Table \ref{tab:memorization}, shows the drop in BLEU scores for each of the models after removing the correct document on OR-QuAC dataset. As can be seen, the percentage drop is highest in the VRAG model indicating a higher usage of knowledge instance and thus, possibly lesser memorization.

\section{Conclusion}
In this paper we described an approach to run variational training on knowledge grounded dialog with a large  corpus. Our experiments on three conversational QA datasets indicate that variational training is helpful as it produces better document samples while training. We find that our model, VRAG (having access to superior samples from posterior while training), not only generates better responses, it also learns a better retriever (prior distribution). 

We believe that such sampling approximations could also be helpful in other tasks; for instance, it could also be interesting to apply them to other approaches such as Reinforcement Learning to the setting of a large corpus. This would require overcoming similar challenges in sampling as we did in this paper for variational training.

\bibliographystyle{named}
\bibliography{ijcai22}

\begin{thebibliography}{}

\bibitem[\protect\citeauthoryear{Baudi{\v{s}} and
  {\v{S}}ediv{\`y}}{2015}]{CuratedTREC}
Petr Baudi{\v{s}} and Jan {\v{S}}ediv{\`y}.
\newblock Modeling of the question answering task in the yodaqa system.
\newblock In {\em International Conference of the Cross-Language Evaluation
  Forum for European Languages}, pages 222--228. Springer, 2015.

\bibitem[\protect\citeauthoryear{Berant \bgroup \em et al.\egroup
  }{2013}]{WebQuestions}
Jonathan Berant, Andrew Chou, Roy Frostig, and Percy Liang.
\newblock Semantic parsing on freebase from question-answer pairs.
\newblock In {\em Proceedings of the 2013 conference on empirical methods in
  natural language processing}, pages 1533--1544, 2013.

\bibitem[\protect\citeauthoryear{Bowman \bgroup \em et al.\egroup
  }{2015}]{bowman2015generating}
Samuel~R Bowman, Luke Vilnis, Oriol Vinyals, Andrew~M Dai, Rafal Jozefowicz,
  and Samy Bengio.
\newblock Generating sentences from a continuous space.
\newblock {\em arXiv preprint arXiv:1511.06349}, 2015.

\bibitem[\protect\citeauthoryear{Campos \bgroup \em et al.\egroup
  }{2020}]{DoQA}
Jon~Ander Campos, Arantxa Otegi, Aitor Soroa, Jan Deriu, Mark Cieliebak, and
  Eneko Agirre.
\newblock {D}o{QA} - accessing domain-specific {FAQ}s via conversational {QA}.
\newblock In {\em Proceedings of the 58th Annual Meeting of the Association for
  Computational Linguistics}, pages 7302--7314, Online, July 2020. Association
  for Computational Linguistics.

\bibitem[\protect\citeauthoryear{Chen \bgroup \em et al.\egroup
  }{2016}]{chen2016variational}
Xi~Chen, Diederik~P Kingma, Tim Salimans, Yan Duan, Prafulla Dhariwal, John
  Schulman, Ilya Sutskever, and Pieter Abbeel.
\newblock Variational lossy autoencoder.
\newblock {\em arXiv preprint arXiv:1611.02731}, 2016.

\bibitem[\protect\citeauthoryear{Chen \bgroup \em et al.\egroup }{2020}]{Che20}
Xiuyi Chen, Fandong Meng, Peng Li, Feilong Chen, Shuang Xu, Bo~Xu, and Jie
  Zhou.
\newblock Bridging the gap between prior and posterior knowledge selection for
  knowledge-grounded dialogue generation.
\newblock In {\em Proceedings of the 2020 Conference on Empirical Methods in
  Natural Language Processing (EMNLP)}, pages 3426--3437, 2020.

\bibitem[\protect\citeauthoryear{Choi \bgroup \em et al.\egroup }{2018}]{QuAC}
Eunsol Choi, He~He, Mohit Iyyer, Mark Yatskar, Wen-tau Yih, Yejin Choi, Percy
  Liang, and Luke Zettlemoyer.
\newblock {Q}u{AC}: Question answering in context.
\newblock In {\em Proceedings of the 2018 Conference on Empirical Methods in
  Natural Language Processing}, pages 2174--2184, Brussels, Belgium,
  October-November 2018. Association for Computational Linguistics.

\bibitem[\protect\citeauthoryear{Devlin \bgroup \em et al.\egroup
  }{2018}]{BERT}
Jacob Devlin, Ming-Wei Chang, Kenton Lee, and Kristina Toutanova.
\newblock Bert: Pre-training of deep bidirectional transformers for language
  understanding.
\newblock {\em arXiv preprint arXiv:1810.04805}, 2018.

\bibitem[\protect\citeauthoryear{Guu \bgroup \em et al.\egroup }{2020}]{Guu20}
Kelvin Guu, Kenton Lee, Zora Tung, Panupong Pasupat, and Ming-Wei Chang.
\newblock Realm: Retrieval-augmented language model pre-training.
\newblock {\em arXiv preprint arXiv:2002.08909}, 2020.

\bibitem[\protect\citeauthoryear{Jang \bgroup \em et al.\egroup
  }{2016}]{Gumbel}
Eric Jang, Shixiang Gu, and Ben Poole.
\newblock Categorical reparameterization with gumbel-softmax.
\newblock {\em arXiv preprint arXiv:1611.01144}, 2016.

\bibitem[\protect\citeauthoryear{Johnson \bgroup \em et al.\egroup
  }{2017}]{JDH17}
Jeff Johnson, Matthijs Douze, and Herv{\'e} J{\'e}gou.
\newblock Billion-scale similarity search with gpus.
\newblock {\em arXiv preprint arXiv:1702.08734}, 2017.

\bibitem[\protect\citeauthoryear{Joshi \bgroup \em et al.\egroup
  }{2017}]{TriviQA}
Mandar Joshi, Eunsol Choi, Daniel~S Weld, and Luke Zettlemoyer.
\newblock Triviaqa: A large scale distantly supervised challenge dataset for
  reading comprehension.
\newblock {\em arXiv preprint arXiv:1705.03551}, 2017.

\bibitem[\protect\citeauthoryear{Karpukhin \bgroup \em et al.\egroup
  }{2020}]{dpr20}
Vladimir Karpukhin, Barlas O{\u{g}}uz, Sewon Min, Ledell Wu, Sergey Edunov,
  Danqi Chen, and Wen-tau Yih.
\newblock Dense passage retrieval for open-domain question answering.
\newblock {\em arXiv preprint arXiv:2004.04906}, 2020.

\bibitem[\protect\citeauthoryear{Kim \bgroup \em et al.\egroup }{2020a}]{KAK20}
Byeongchang Kim, Jaewoo Ahn, and Gunhee Kim.
\newblock Sequential latent knowledge selection for knowledge-grounded
  dialogue.
\newblock {\em arXiv preprint arXiv:2002.07510}, 2020.

\bibitem[\protect\citeauthoryear{Kim \bgroup \em et al.\egroup }{2020b}]{DSTC9}
Seokhwan Kim, Mihail Eric, Karthik Gopalakrishnan, Behnam Hedayatnia, Yang Liu,
  and Dilek Hakkani-Tur.
\newblock Beyond domain {API}s: Task-oriented conversational modeling with
  unstructured knowledge access.
\newblock In {\em Proceedings of the 21th Annual Meeting of SIGDIAL}, pages
  278--289, 1st virtual meeting, July 2020. Association for Computational
  Linguistics.

\bibitem[\protect\citeauthoryear{Kingma and Welling}{2013}]{KW13}
Diederik~P Kingma and Max Welling.
\newblock Auto-encoding variational bayes.
\newblock {\em arXiv preprint arXiv:1312.6114}, 2013.

\bibitem[\protect\citeauthoryear{Kwiatkowski \bgroup \em et al.\egroup
  }{2019}]{NQ}
Tom Kwiatkowski, Jennimaria Palomaki, Olivia Redfield, Michael Collins, Ankur
  Parikh, Chris Alberti, Danielle Epstein, Illia Polosukhin, Jacob Devlin,
  Kenton Lee, et~al.
\newblock Natural questions: a benchmark for question answering research.
\newblock {\em Transactions of the Association for Computational Linguistics},
  7:453--466, 2019.

\bibitem[\protect\citeauthoryear{Lewis \bgroup \em et al.\egroup }{2020}]{RAG}
Patrick Lewis, Ethan Perez, Aleksandra Piktus, Fabio Petroni, Vladimir
  Karpukhin, Naman Goyal, Heinrich K{\"u}ttler, Mike Lewis, Wen-tau Yih, Tim
  Rockt{\"a}schel, et~al.
\newblock Retrieval-augmented generation for knowledge-intensive nlp tasks.
\newblock {\em arXiv preprint arXiv:2005.11401}, 2020.

\bibitem[\protect\citeauthoryear{Li \bgroup \em et al.\egroup
  }{2020}]{neurips_lin}
Linxiao Li, Can Xu, Wei Wu, YUFAN ZHAO, Xueliang Zhao, and Chongyang Tao.
\newblock Zero-resource knowledge-grounded dialogue generation.
\newblock In H.~Larochelle, M.~Ranzato, R.~Hadsell, M.~F. Balcan, and H.~Lin,
  editors, {\em Advances in Neural Information Processing Systems}, volume~33,
  pages 8475--8485. Curran Associates, Inc., 2020.

\bibitem[\protect\citeauthoryear{Lian \bgroup \em et al.\egroup }{2019}]{Lia19}
Rongzhong Lian, Min Xie, Fan Wang, Jinhua Peng, and Hua Wu.
\newblock Learning to select knowledge for response generation in dialog
  systems.
\newblock {\em arXiv preprint arXiv:1902.04911}, 2019.

\bibitem[\protect\citeauthoryear{Loshchilov and Hutter}{2017}]{AdamW}
Ilya Loshchilov and Frank Hutter.
\newblock Decoupled weight decay regularization.
\newblock {\em arXiv preprint arXiv:1711.05101}, 2017.

\bibitem[\protect\citeauthoryear{Lucas \bgroup \em et al.\egroup
  }{2019}]{lucas2019understanding}
James Lucas, George Tucker, Roger Grosse, and Mohammad Norouzi.
\newblock Understanding posterior collapse in generative latent variable
  models.
\newblock {\em ICLR 2019 Workshop DeepGenStruct}, 2019.

\bibitem[\protect\citeauthoryear{Maddison \bgroup \em et al.\egroup
  }{2016}]{Concrete}
Chris~J Maddison, Andriy Mnih, and Yee~Whye Teh.
\newblock The concrete distribution: A continuous relaxation of discrete random
  variables.
\newblock {\em arXiv preprint arXiv:1611.00712}, 2016.

\bibitem[\protect\citeauthoryear{Oord \bgroup \em et al.\egroup
  }{2017}]{oord2017neural}
Aaron van~den Oord, Oriol Vinyals, and Koray Kavukcuoglu.
\newblock Neural discrete representation learning.
\newblock {\em arXiv preprint arXiv:1711.00937}, 2017.

\bibitem[\protect\citeauthoryear{Paranjape \bgroup \em et al.\egroup
  }{2022}]{paranjape2022hindsight}
Ashwin Paranjape, Omar Khattab, Christopher Potts, Matei Zaharia, and
  Christopher~D Manning.
\newblock Hindsight: Posterior-guided training of retrievers for improved
  open-ended generation.
\newblock In {\em International Conference on Learning Representations}, 2022.

\bibitem[\protect\citeauthoryear{Qu \bgroup \em et al.\egroup
  }{2020}]{OpenQuAC}
Chen Qu, Liu Yang, Cen Chen, Minghui Qiu, W.~Bruce Croft, and Mohit Iyyer.
\newblock {\em Open-Retrieval Conversational Question Answering}, page
  539–548.
\newblock Association for Computing Machinery, New York, NY, USA, 2020.

\bibitem[\protect\citeauthoryear{Radford \bgroup \em et al.\egroup
  }{2019}]{GPT2}
Alec Radford, Jeffrey Wu, Rewon Child, David Luan, Dario Amodei, and Ilya
  Sutskever.
\newblock Language models are unsupervised multitask learners.
\newblock {\em OpenAI blog}, 1(8):9, 2019.

\bibitem[\protect\citeauthoryear{Robertson \bgroup \em et al.\egroup
  }{1994}]{BM25}
Stephen~E. Robertson, Steve Walker, Susan Jones, Micheline Hancock-Beaulieu,
  and Mike Gatford.
\newblock Okapi at trec-3.
\newblock In {\em TREC}, 1994.

\end{thebibliography}

\appendix

\section{Hyperparameters and Training Details}
\label{sec:hyper}
\subsection{Architecture }
We now describe the neural networks used to model the prior and posterior distributions, and the response generator. 
\subsubsection{Prior Distribution}
We need to create document and context representations defined by functions $f$and $g$, respectively for modelling the prior. 

To define the function $g$ we use the BERT Model \cite{BERT}. We consider the context $\bx$ as a sequence of turns $x_1, \dots x_i, \dots x_n$, where each turn $x_i$, is an utterance from either one of the two speakers `S1' or `S2' in the dialog. We then create an input representation of the context 
= $[CLS]$, \textlangle $S1$\textrangle, $x_1$, \textlangle $S2$\textrangle, $x_2$, \textlangle $S1$\textrangle, ..., $x_n$, $[SEP]$, where $[CLS]$, $[SEP]$, \textlangle $S1$\textrangle, \textlangle $S2$ \textrangle ~are marker tokens.
This input representation is then fed to BERT;  the embedding at the final layer of $[CLS]$ token is then passed through a linear layer to obtain the embedding representation $g(\bx)$ of the context $\bx$.

Similarly enc($\bz$) = $[CLS]$, $\bz$, $[SEP]$ forms the input encoding for document $\bz$. This is also passed through a (separate) BERT model,  followed by the application of a linear layer to the embedding at $[CLS]$ token. This gives the embedding representation $f(\bz)$ for document $\bz$.

\subsubsection{Posterior Distribution}
Similar to the modeling of the prior distribution, we use another (separate) BERT to model the posterior. Here we create the input representation of the context-response  pair ($\bx$, $\by$) 
as: 
$[CLS]$, \textlangle $S1$\textrangle, $x_1$, \textlangle $S2$\textrangle, $x_2$, \textlangle $S1$\textrangle, ..., $x_n$, \textlangle $RSEP$\textrangle, $\by$, $[SEP]$. Here, \textlangle $RSEP$\textrangle is a special marker token used to separate the response from the dialog-context. This input representation is passed through BERT model and we obtain the representation $g(\bx, \by)$ by applying a linear layer on the embedding of the $[CLS]$ token at last layer.
\subsubsection{Response Likelihood}
We use the standard sequence to sequence formulation for response likelihood, where
\begin{equation}
    \log p(\by\vert \bz,\bx) = \sum_j \log p(\by_j\vert \by_{<j},\bz,\bx).
\end{equation}
Here, we use the GPT2~\cite{GPT2} model as our decoder. We encode the input sequence as \textlangle $bos$\textrangle, enc($\bx$), \textlangle $KSEP$\textrangle, enc($\bz$), \textlangle $eos$\textrangle, where \textlangle $bos$\textrangle ~and \textlangle $eos$\textrangle ~are the begin- and end-of-sequence markers, and \textlangle $KSEP$\textrangle ~is a special marker token to separate the dialog context from the document (knowledge).

For each response token index $j$, $\log p(\by_j \vert \by_{<j},\bz,\bx)$ is then defined by passing the output at the $j^{\text{th}}$ index of the decoder through a linear layer.
\subsection{Network Hyperparameters}

We train both RAG and VRAG with AdamW optimizer. We train for 10 epochs with early stopping (patience = 5). The learning rates and time per epoch are given in Table \ref{table:training_details}. We use BERT base (110 million parameters) for the context encoders in both prior and posterior networks and GPT2 small (117 million parameters) for the decoder network. The document encoder is also a BERT base (110 million parameters) which is fixed at training time.
All of our experiments are conducted on a single machine with 100GB of system memory and Nvidia V100 GPU.

\section{Further discussion of experiments and results}
\subsection{Dataset Statistics}
\label{sec:dataset}
The statistics of datasets we used are shown in Table \ref{table:datasets_statistics}.

\subsection{Effect of Decoder fine-tuning}
Training VRAG involves optimizing two objectives - reducing the KL-divergence between the document-prior and document-posterior, and, maximizing the log likelihood of the responses. VAE  models often end up prioritizing the KL-divergence over the likelihood objective and sometimes end up with zero KL-divergence by forcing the document-posterior to match the prior (called posterior-collapse) \cite{lucas2019understanding,bowman2015generating,chen2016variational,oord2017neural}. However, we hypothesize even in cases where there is no posterior collapse, the joint training could result in the response-generator (likelihood term) being inadequately trained.

We observed that fine-tuning the response decoder after freezing the weights of priors, results in an improvement in both RAG and VRAG.
The results shown in the main paper were obtained after fine tuning the generator after the joint training is complete.
We show the results both with and without fine tuning in Table \ref{table:all} here.
It is interesting to note on the DoQA dataset, RAG initially reported a B-4 of only $0.69$ (top-5 decoding) which goes up to $17.04$ after fine-tuning .
\subsection{Effect of top-k}
\label{sec:Effect of top-k}

\begin{figure}[t]
\centering
\includegraphics[width=0.4\textwidth]{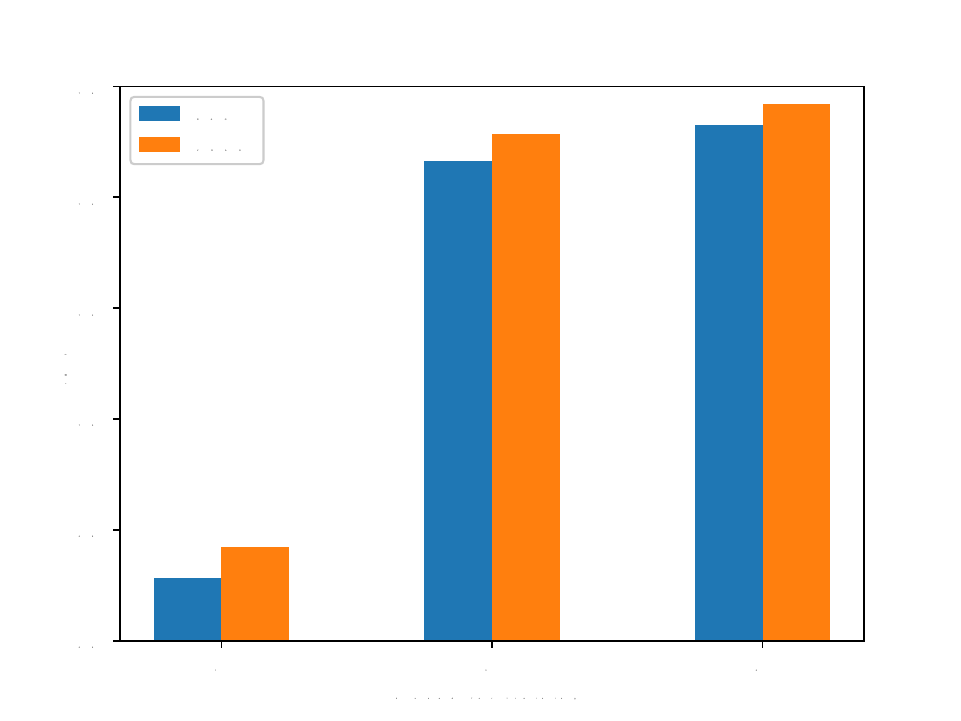}
\caption{Variation in Recall as models are trained with different values of $k$ (top-k documents). Results shown with k = 3 on the DoQA dataset.}
\label{fig:recall}
\end{figure}

\begin{table}[H]
\begin{center}
\begin{tabular}{|c|c|c|c|} 
\hline
& OR-QuAC & DSTC9 & DoQA \\
\hline
Train instances & 25,942 & 19,184 & 3,341 \\
Val. instances & 2,828 & 2,673 & 662 \\
Test instances & 4,421 & 1,981 & 1,263 \\
Knowledge documents & 68,031 & 12,039 & 1,108 \\
 \hline
\end{tabular}
\end{center}
\caption{Dataset statistics : Each `instance' refers to a dialog context-response pair. }
\label{table:datasets_statistics}
\end{table}

We also consider the case when we reduce the number of samples ($k$) while training. Figure \ref{fig:recall} shows the variation in $R@3$ versus $k$ on the DoQA dataset. Here we observe that recall decreases with decrease in $k$. This is because the computation of objective becomes less accurate for decrease in $k$. We also note that when only one document is sampled, RAG outperforms VRAG. We hypothesize, that in this case the VRAG prior trains poorly due to an extremely weak approximation of the KL divergence\footnote{approximation has just 1-2 samples when k=1}.

\subsection {Recall scores versus training epoch}
\label{sec:epoch}
\begin{figure*}[h!]
    \centering
    \begin{subfigure}[b]{0.3\textwidth}
        \includegraphics[width=\textwidth]{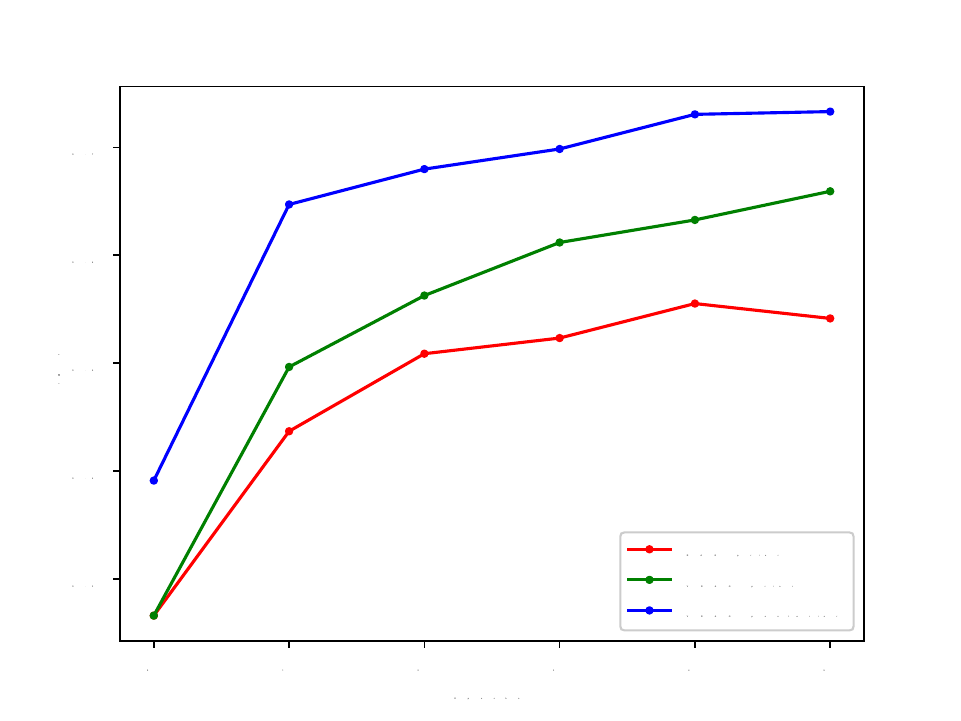}
    \caption{Recall@1}
    \end{subfigure}
    \hfill
    \begin{subfigure}[b]{0.3\textwidth}
        \includegraphics[width=\textwidth]{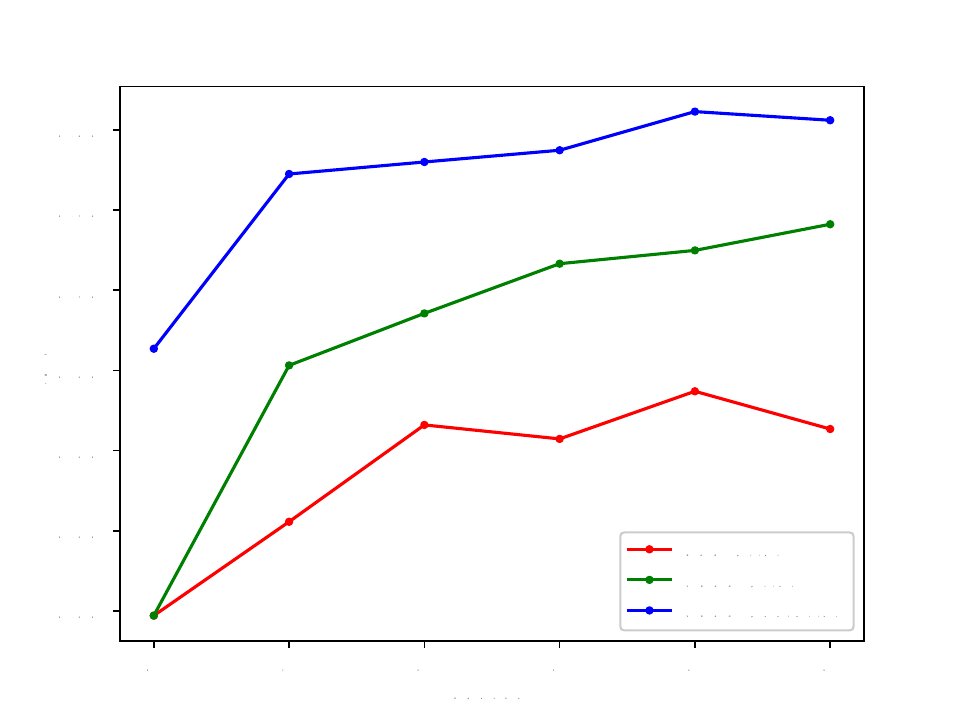}
     \caption{Recall@5}
    \end{subfigure}
    \hfill
    \begin{subfigure}[b]{0.3\textwidth}
        \includegraphics[width=\textwidth]{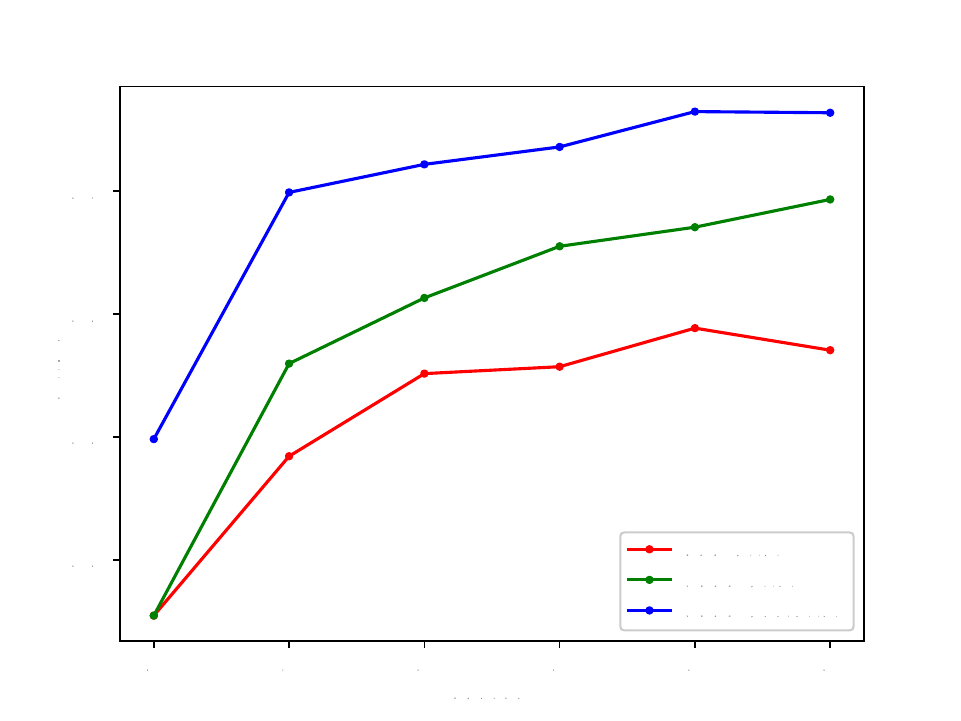}
     \caption{MRR@5}
    \end{subfigure}
    \caption{Recall versus epoch numbers on training set for DoQA dataset for RAG prior, VRAG prior and posterior distributions.}
    \label{fig:recall-plot}
\end{figure*}

As shown in Figure \ref{fig:recall-plot}, the recall scores improve over epoch. Moreover the posterior recall scores are always higher then prior recall scores. This shows both the benefit of using a posterior during training, as well as the benefit of training the retriever as opposed to keeping it fixed.

\subsection{RAG prior vs VRAG posterior}
\label{sec:posterior}
Table \ref{table:recall} compares the recall of RAG prior with VRAG posterior. The VRAG (posterior scores) being higher than RAG (prior) help in generating better document samples for the response generator during training. As discussed in the main section this explains the improved scores of VRAG over RAG.

\section{Dialog examples}
\label{sec:Dialog examples}

\begin{table*}[h!]
\centering
\footnotesize
\begin{tabular}{|c|c|c|c|c|c|c|c|c|}
\hline
& \multicolumn{2}{c|}{OR-QuAC} & \multicolumn{2}{c|}{DSTC9} & \multicolumn{2}{c|}{DoQA} & \multicolumn{2}{c|}{OR-QuAC-QA}
\\
\hline
& $R@1$ & $R@5$ & $R@1$ & $R@5$ & $R@1$ & $R@5$ & $R@1$ & $R@5$
\\
\hline
RAG (Prior) & 24.2 & 54.9 & 69.4 & 84.3 & 61.8 & 79.0 & 15.0 & 38.1
\\
VRAG (Posterior) & 39.0 & 72.4 & 82.4 & 90.6 & 82.7 & 93.2 & 31.6 & 65.4
\\
\hline
\end{tabular}
\caption{Recall scores for RAG (Prior) and VRAG (Posterior)}
\label{table:recall}
\end{table*}

\begin{table*}[h!tbp]
\begin{center}
\begin{tabular}{|c|c|c|c|c|c|c|c|c|c|}
\hline
& & \multicolumn{4}{c|}{top-1 decoding} & \multicolumn{4}{c|}{top-5 decoding}
\\\hline
{\bf Dataset} & {\bf Model} & {\bf B-1} & {\bf B-4} & {\bf BP-1} & {\bf BP-4} & {\bf B-1} & {\bf B-4} & {\bf BP-1} & {\bf BP-4}
\\
\hline
& DPR + GPT2 & 13.65 & 6.11 & 4.41 & 3.11 & 16.06 & 7.97 & 11.36 & 7.37
\\
& RAG & 13.06 & 6.19 & 4.68 & 3.19 & 15.73 & 7.87 & 12.09 & 7.49
\\
& pre-VM & 9.49 & 3.58 & 2.85 & 1.84 & 10.87 & 4.32 & 6.91 & 4.02
\\
OR-QuAC & VRAG & 11.80 & 5.83 & 4.73 & 3.26 & 13.54 & 6.71 & 10.76 & 6.41
\\
\cite{OpenQuAC} & RAG (fine-tuned) & 12.88 & 5.94 & 4.60 & 3.03 & 15.39 & 7.64 & 11.72 & 7.21
\\
& pre-VM (fine-tuned) & 11.44 & 4.87 & 3.52 & 2.36 & 13.55 & 6.26 & 9.17 & 5.89
\\
& VRAG (fine-tuned) & 13.97 & 7.58 & 5.61 & 4.02 & 16.30 & 9.11 & 13.10 & 8.72
\\
\hline
& DPR + GPT2 & 31.84 & 7.21 & 4.37 & 1.08 & 31.81 & 7.17 & 11.14 & 2.60
\\
& RAG & 32.54 & 8.35 & 24.47 & 6.89 & 32.55 & 8.33 & 28.17 & 7.43
\\
& pre-VM & 24.63 & 4.19 & 7.22 & 1.64 & 25.09 & 4.33 & 10.95 & 2.29
\\
DSTC9 & VRAG & 32.76 & 8.60 & 25.81 & 7.36 & 32.79 & 8.64 & 29.18 & 7.90
\\
\cite{DSTC9} & RAG (fine-tuned) & 33.28 & 8.26 & 25.87 & 6.86 & 33.30 & 8.27 & 28.75 & 7.45
\\
& pre-VM (fine-tuned) & 31.57 & 7.16 & 9.92 & 2.66 & 31.87 & 7.29 & 14.7 & 3.98
\\
& VRAG (fine-tuned) & 33.49 & 8.70 & 26.49 & 7.57 & 33.51 & 8.67 & 29.80 & 8.03
\\
\hline
& DPR + GPT2 & 21.26 & 14.31 & 17.83 & 14.20 & 22.60 & 15.73 & 20.53 & 15.62
\\
& RAG & 2.19 & 0.69 & 1.73 & 0.64 & 3.23 & 0.77 & 2.68 & 0.74
\\
& pre-VM & 4.27 & 1.26 & 3.21 & 1.21 & 5.32 & 1.30 & 4.61 & 1.25
\\
DoQA & VRAG & 17.54 & 11.70 & 15.52 & 11.70 & 17.87 & 11.85 & 16.73 & 11.85
\\
\cite{DoQA} & RAG (fine-tuned) & 23.59 & 17.04 & 20.86 & 16.92 & 24.27 & 17.73 & 22.75 & 17.60
\\
& pre-VM (fine-tuned) & 23.33 & 16.70 & 20.47 & 16.5 & 23.79 & 17.21 & 22.24 & 17.07
\\
& VRAG (fine-tuned) & 23.38 & 17.02 & 20.91 & 16.94 & 23.29 & 16.88 & 21.93 & 16.80
\\
\hline
& DPR + GPT2 & 9.11 & 1.88 & 1.81 & 1.07 & 10.18 & 2.61 & 4.95 & 2.38
\\
& RAG & 9.30 & 2.49 & 2.61 & 1.56 & 10.81 & 3.35 & 6.36 & 3.17
\\
& pre-VM & 5.99 & 0.83 & 0.68 & 0.32 & 6.84 & 1.07 & 2.28 & 1.01
\\
OR-QuAC-QA & VRAG & 9.16 & 2.76 & 2.88 & 1.73 & 10.46 & 3.25 & 6.60 & 3.16
\\
\cite{OpenQuAC} & RAG (fine-tuned) & 9.12 & 2.31 & 2.36 & 1.44 & 10.39 & 2.97 & 5.75 & 2.75
\\
& pre-VM (fine-tuned) & 6.96 & 1.17 & 1.11 & 0.67 & 7.84 & 1.75 & 3.05 & 1.67
\\
& VRAG (fine-tuned) & 9.64 & 2.93 & 2.83 & 1.66 & 10.65 & 3.49 & 6.73 & 3.36
\\
\hline
\end{tabular}
\caption{Table showing complete results both with and without fine tuning the models}
\label{table:all_full}
\end{center}
\end{table*}

\begin{table}[t]
\footnotesize
\begin{center}
\begin{tabular}{|c|c|c|} 
\hline
Dataset & Learning rate & Time per epoch\\
\hline
OR-QuAC & $10^{-5}$ & 3 hrs\\
DSTC9 & $10^{-6}$ & 2 hrs\\
DoQA & $10^{-6}$ & 20 mins\\
OR-QuAC-QA & $10^{-5}$ & 2.5 hrs\\
\hline
\end{tabular}
\end{center}
\caption{Hyperparameters and training time for each epoch}
\label{table:training_details}
\end{table}

Here, we show the retrieved documents for a given dialog context and the generated responses using top-1 decoding (using best document). All documents in the tables are in decreasing order of probabilities.

\subsection{OR-QuAC}
Refer to table \ref{table:OR-QuAC} for examples.

\begin{table}[H]
\begin{flushleft}
\textbf{Dialog context}
\\
\textcolor{orange}{\textbf{User}: What disputes did Billy Graham have?
\\
\textbf{Agent}: Graham personally sued Zahorian and the WWF.
\\
\textbf{User}: Why?
\\
\textbf{Agent}: He claimed that they had forced him to take steroids to maintain his position in the company.
\\
\textbf{User}: Did he win the lawsuit?
\\
\textbf{Agent}: His lawsuit was unsuccessful.
\\
\textbf{User}: What happened after the suit failed?}
\\
\textbf{Retrieved documents}
\\
\end{flushleft}
\centering
\begin{tabular}{|l|l|}
\hline
\multicolumn{1}{|c|}{RAG} & \multicolumn{1}{|c|}{VRAG}
\\
\hline
Late in 2003, Giambi was & \textcolor{Emerald}{In the early 1990s}
\\
named by FBI officers & \textcolor{Emerald}{US federal agents were}
\\
investigating the Bay Area ... & \textcolor{Emerald}{investigating Dr. Geo ...}
\\
\hline
Landis testified at the & Late in 2003, Giambi
\\
hearing that Geoghegan & was named by FBI 
\\
came to know of LeMond's ... & officers investigating ...
\\
\hline
\textcolor{Emerald}{In the early 1990s US} & Since 2003, Bonds has
\\
\textcolor{Emerald}{federal agents were} & been a key figure in the
\\
\textcolor{Emerald}{investigating Dr. George ...} & Bay Area Laboratory ...
\\
\hline
\end{tabular}
\caption{Documents retrieved by RAG and VRAG (in decreasing order of probabilities). The green highlighted document is the correct document required to generate the response.}
\begin{flushleft}
\textbf{Response}
\\
\textcolor{Emerald}{\textbf{Ground truth response}: Graham went on a public awareness campaign regarding the dangers of steroids.}
\\
\textcolor{BrickRed}{\textbf{Response (RAG)}: The prosecution in the Barry Bonds perjury case indicated they intended to call both Jason and Jeremy Giambi to testify against Bonds in his March 2009 trial.}
\\
\textcolor{Emerald}{\textbf{Response (VRAG)}: Graham went on a public awareness campaign regarding the dangers of steroids during this time, including an appearance with McMahon on The Phil Donahue Show in 1992.}
\end{flushleft}
\label{table:OR-QuAC}
\end{table}

\subsection{DSTC9}
Since a lot of entities contain similar documents in the DSTC9 \cite{DSTC9} dataset, we create a document collection in which the documents are represented as 
\begin{equation}
\text{ENTITY\_NAME : question : answer}
\end{equation}
to distinguish between the documents of different entities. Refer to table \ref{table:DSTC9} for examples.

\subsection{DoQA}
Both models, RAG and VRAG generate the correct response as shown in the example, however, in this example, RAG does so by using the incorrect document which shows response memorization. Refer to table \ref{table:DoQA} for examples.

\begin{table}[h!]
\begin{flushleft}
\textbf{Dialog context}
\\
\textcolor{orange}{\textbf{User}: Hi. I'm trying to find a cheap place to eat that serves Izakaya food in Laurel Heights.
\\
\textbf{Agent}: There isn't a cheap Izakaya place in Laurel Heights. Would you like for me to check a different price range?
\\
\textbf{User}: Moderate price range should be fine.
\\
\textbf{Agent}: I found the Kaiju Eats Ramen and Izakaya in Laurel Heights. It is moderately priced.
\\
\textbf{User}: Do they take advanced reservations?}
\\
\textbf{Retrieved documents}
\\
\end{flushleft}
\centering
\begin{tabular}{|l|l|}
\hline
\multicolumn{1}{|c|}{RAG} & \multicolumn{1}{|c|}{VRAG}
\\
\hline
KAIJU EATS RAMEN & \textcolor{Emerald}{KAIJU EATS RAMEN}
\\
AND IZAKYA : Can I order & \textcolor{Emerald}{AND IZAKYA : Can I}
\\
food for delivery at Kaiju & \textcolor{Emerald}{make an advance}
\\
Eats Ramen and Izakaya? : & \textcolor{Emerald}{reservation? : Yes,}
\\
Delivery service is not & \textcolor{Emerald}{reservations taken.}
\\
available at Kaiju Eats &
\\
Ramen and Izakaya. &
\\
\hline
MENSHO TOKYO : Are & KAIJU EATS RAMEN
\\
reservations necessary there? & AND IZAKYA : Can I
\\
: No, Mensho Tokyo doesn't & order food for delivery at
\\
accept reservations. & Kaiju Eats Ramen and
\\
& Izakaya? : Delivery
\\
& service is not available
\\
& at Kaiju Eats
\\
& Ramen and Izakaya.
\\
\hline
TABOUN : Do you accept & USHIO RAMEN :
\\
reservations? : No, they do & Do you accept
\\
not take reservations. & reservations? : Yes,
\\
& the restaurant does
\\
& take reservations.
\\
\hline
\end{tabular}
\caption{Documents retrieved by RAG and VRAG (in decreasing order of probabilities). The green highlighted document is the correct document required to generate the response. The text in CAPITALS is the name of the entity.}
\begin{flushleft}
\textbf{Response}
\\
\textcolor{Emerald}{\textbf{Ground truth response}: Yes, they do take reservations. Would you like me to make one for you?}
\\
\textcolor{BrickRed}{\textbf{Response (RAG)}: No, they do not take advanced reservations. Anything else I can do for you?}
\\
\textcolor{Emerald}{\textbf{Response (VRAG)}: Yes, you may make a reservation. Would you like to make one?}
\end{flushleft}
\label{table:DSTC9}
\end{table}

\begin{table}[h!]
\begin{flushleft}
\textbf{Dialog context}
\\
\textcolor{orange}{\textbf{User}: Should chicken be cooked beforehand while making pizza?
\\
\textbf{Agent}: Cook the chicken ahead of time.
\\
\textbf{User}: Why should it be cooked ahead of time?}
\\
\textbf{Retrieved documents}
\\
\end{flushleft}
\centering
\begin{tabular}{|l|l|}
\hline
\multicolumn{1}{|c|}{RAG} & \multicolumn{1}{|c|}{VRAG}
\\
\hline
I'd suggest skinless & \textcolor{Emerald}{Cook the chicken ahead}
\\
bone-in chicken thighs, & \textcolor{Emerald}{of time. I doubt your}
\\
as they have plenty of fat & \textcolor{Emerald}{pizza cooking time}
\\
and collagen to keep them & \textcolor{Emerald}{and your chicken}
\\
moist and tasty. ... & \textcolor{Emerald}{cooking time will ...}
\\
\hline
\textcolor{Emerald}{Cook the chicken ahead} & I'd say it depends. If
\\
\textcolor{Emerald}{of time. I doubt your} & you are dead-set on
\\
\textcolor{Emerald}{pizza cooking time} & observing all food
\\
\textcolor{Emerald}{and your chicken} & safety rules, the pizzas
\\
\textcolor{Emerald}{cooking time will ...} & are probably unsafe
\\
& to eat. ...
\\
\hline
Don't cook the chicken & If your chicken burns, the
\\
pieces for so long. Add & issue is likely not that you
\\
them ten or twenty minutes & cooked it too long, but
\\
before serving. For that & that you cooked it too ...
\\
matter, I wouldn't ... &
\\
\hline
\end{tabular}
\caption{Documents retrieved by RAG and VRAG (in decreasing order of probabilities). The green highlighted document is the correct document required to generate the response.}
\begin{flushleft}
\textbf{Response}
\\
\textcolor{Emerald}{\textbf{Ground truth response}: I doubt your pizza cooking time and your chicken cooking time will be a perfect match, and it is more likely than anything that your chicken would be undercooked. That would not be good for anyone.}
\\
\textcolor{Emerald}{\textbf{response (RAG)}: I doubt your pizza cooking time and your chicken cooking time will be a perfect match, and it is more likely than anything that your chicken would be undercooked.}
\\
\textcolor{Emerald}{\textbf{response (VRAG)}: I doubt your pizza cooking time and your chicken cooking time will be a perfect match, and it is more likely than anything that your chicken would be undercooked.}
\end{flushleft}
\label{table:DoQA}
\end{table}

\end{document}